\documentclass[a4paper, 10pt, conference]{ieeeconf}
\usepackage{FG2026}
\FGfinalcopy

\usepackage{cite}
\usepackage{amsmath,amssymb,amsfonts}
\usepackage{graphicx}
\usepackage{textcomp}
\usepackage{xcolor}
\usepackage{xspace}
\usepackage{booktabs}
\usepackage{multirow}
\usepackage{url}
\usepackage[hidelinks]{hyperref}
\usepackage{microtype}
\usepackage{url}
\usepackage{microtype}
\usepackage{balance}
\usepackage{pifont}
\usepackage[T1]{fontenc}
\usepackage{xspace} 
\newcommand{\mypartitle}[2][2.7]{\vspace*{-#1 ex}~\\{\noindent {\bf #2}}}
\newcommand{\fvv}{Face-vid2vid}
\newcommand{\nvfair}{\textsc{nvfair}}

\newcommand{\pdiff}{\textsc{pixel\_diff}}
\newcommand{\rawfeat}{\textsc{raw\_feat}}
\newcommand{\static}{\textsc{static}}
\newcommand{\fcc}{\textsc{fcc}}
\newcommand{\ccc}{\textsc{ccc}}
\newcommand{\yes}{\ding{51}}
\newcommand{\no}{\ding{55}}
\newcommand{\Video}{\ensuremath{V}\xspace}
\newcommand{\VideoID}[2]{\ensuremath{V_{#1}^{#2}}\xspace}
\newcommand{\featdiff}{\textsc{feat\_diff}}
\newcommand{\rndiff}{\textsc{resnet18-feat\_diff}}
\newcommand{\rnpdiff}{\textsc{resnet18-pixel\_diff}}
\newcommand{\rnraw}{\textsc{resnet18-raw\_feat}}
\newcommand{\rnstatic}{\textsc{resnet18-static}}
\def\FGPaperID{441}

\title{Micro-Expression-Aware Avatar Fingerprinting via\\
Inter-Frame Feature Differencing}

\author{\parbox{16cm}{\centering
    {\large Masoumeh Chapariniya$^{1,2}$, Jean-Marc Odobez$^{3,4}$,
             Volker Dellwo$^{1}$, Teodora Vukovi\'{c}$^{1,2}$}\\[4pt]
    {\normalsize $^1$Department of Computational Linguistics, University of Zurich, Zurich, Switzerland}\\
    {\normalsize $^2$UZH Digital Society Initiative}\\
    {\normalsize $^3$Idiap Research Institute, Martigny, Switzerland}\\
    {\normalsize $^4$\'{E}cole Polytechnique F\'{e}d\'{e}rale de Lausanne (EPFL), Lausanne, Switzerland}\\[2pt]
    {\small \texttt{masoumeh.chapariniya@uzh.ch},\ \texttt{odobez@idiap.ch},\
             \texttt{volker.dellwo@uzh.ch},\ \texttt{teodora.vukovic2@uzh.ch}}
}}

\usepackage{fancyhdr}

\begin{document}

\ifFGfinal
  \thispagestyle{fancy}\pagestyle{empty}
\else
  \author{Anonymous FG2026 submission\\ Paper ID \FGPaperID}
  \pagestyle{plain}
\fi

\maketitle
\thispagestyle{fancy}
\let\thefootnote\relax
\footnotetext{Code: \href{https://github.com/Mchapariniya/TrustFA-FG2026}{https://github.com/Mchapariniya/TrustFA-FG2026}}

\begin{abstract}
Avatar fingerprinting, i.e., verifying \emph{who drives} a synthetic talking-head video rather than whether it is real, is a critical safeguard for authorized use of face-reenactment technology.
Existing methods rely on a fixed, non-differentiable landmark extraction stage that prevents the fingerprinting model from being optimized end-to-end from raw pixels.
We propose a preprocessing-free system built on a micro-expression-aware backbone operating on raw video frames, with \textbf{inter-frame feature differencing} as the core design principle: consecutive feature maps are subtracted in the learned deep feature space, so that temporally stable appearance dimensions contribute zero to the output while driver-specific motion dynamics are preserved.
A controlled ablation on \nvfair{} confirms that temporal motion accounts for the large majority of discriminative performance, and that raw appearance features actively degrade identity separation.
Both the choice of backbone and the differencing principle are essential: differencing alone is insufficient when applied to a generic encoder, as appearance-dominated features collapse to near-identical representations across adjacent frames, while the micro-expression-aware F5C backbone retains measurable motion variation that the differencing operation can exploit.
Without any external preprocessing, our model achieves an overall AUC of \textbf{0.877} on \nvfair{} and matches or exceeds the landmark-based baseline on the majority of cross-generator pairs. 
\end{abstract}

\vspace{0.5em}\noindent\textbf{Keywords:}
avatar fingerprinting, face reenactment, inter-frame feature differencing,
temporal facial dynamics, appearance-agnostic representation,
supervised contrastive learning,
cross-generator generalization.
\vspace{0.5em}
\section{Introduction}
\label{sec:intro}
The rapid proliferation of neural talking-head generators has
fundamentally changed the threat landscape for digital identity.
Keypoint-based generators such as \fvv{}~\cite{wang2021facevidvid}
and TPS~\cite{zhao2022tps}, latent-space systems such as
LIA~\cite{wang2023lia} and LivePortrait~\cite{guo2024liveportrait},
and the newest diffusion-based models~\cite{tian2024emo}
can synthesize photorealistic talking-head videos from a single
image in real time.
While these technologies enable immersive telepresence and
accessibility applications, they simultaneously lower the barrier
for \textbf{unauthorized avatar impersonation}: an attacker can
drive another person's digital likeness without consent, at scale,
using only a single photograph.
Recent surveys identify AI-generated impersonation as a primary
emerging threat to identity verification
systems~\cite{identity2025threats}, and regulatory frameworks such
as the EU AI Act explicitly require provenance mechanisms for
synthetic face media.

Avatar impersonation presents a forensic challenge that is
fundamentally different from standard deepfake detection.
In deepfake detection, the goal is to distinguish real from
synthetic content, where artifact-based
detectors~\cite{haliassos2021lips} exploit rendering
artifacts that remain consistent within a given generator.
Avatar videos, however, require a different question:
\emph{who drives the avatar?}
All inputs are legitimate: the target face image is genuine and the generator is authorized. Only the \emph{driving identity} is unauthorized.
Standard deepfake detectors fail completely here because
generator artifacts are identical regardless of who drives the avatar.
\emph{Avatar fingerprinting}~\cite{prashnani2024avatar} addresses
this by extracting a biometric signature of the driving identity
from the avatar's temporal facial dynamics.
The core challenge is that visual appearance is identical across
all drivers of the same target face.
The only discriminative signal is \emph{temporal facial motion},
the idiosyncratic micro-dynamics each person imprints during
animation~\cite{hill2001categorizing,knappmeyer2001facial,otoole2002recognizing}.
Existing avatar fingerprinting
methods~\cite{prashnani2024avatar,pedrouzo2026avaprintdb,sahadewa2026biasaverse} extract these dynamics by first running
a facial landmark detector on each frame, then training a
downstream model on the resulting geometric features.
While effective, this design imposes a structural constraint:
the landmark stage is frozen and non-differentiable, so its
feature vocabulary is fixed independently of what the
fingerprinting objective actually requires.

Sub-landmark facial dynamics likely carry identity information that is too fine-grained to be reliably captured by landmark-based pipelines and too easily suppressed by generic appearance-oriented encoders. We therefore introduce a preprocessing-free avatar fingerprinting framework that combines a micro-expression-aware F5C backbone with inter-frame feature differencing. The backbone preserves coordinated fine-scale facial micro-expression directly from raw frames, whereas differencing removes temporally stable target appearance and retains only temporal variation. This combination is central to our approach: the backbone produces motion-sensitive features, and the differencing step transforms them into an appearance-agnostic representation of driving identity.

Our contributions are as follows:
\begin{enumerate}
\item We propose a preprocessing-free, end-to-end avatar fingerprinting framework that combines a micro-expression-aware F5C backbone with inter-frame feature differencing.

\item Through controlled ablations, we show that temporal differencing is the main source of identity signal, raw appearance features are harmful, generic backbones are poorly suited to feature-space differencing, and longer clips improve performance.

\item Our 0.53\,M-parameter model achieves 0.877 overall AUC on \nvfair{}, matches or exceeds the landmark-based baseline on most cross-generator pairs, and eliminates external landmark tracking.
\end{enumerate}

\section{Related Work}
\label{sec:related}

\subsection{Avatar Fingerprinting}

Avatar fingerprinting was introduced by Prashnani et al.~\cite{prashnani2024avatar}, who formulated the task as recovering driving identity from temporal facial dynamics in synthetic talking-head videos. Their method represents each frame using facial landmarks, converts them into pairwise normalized Euclidean distances, and learns driver-specific embeddings with a temporal CNN trained using a time-shuffling contrastive objective. Importantly, their ablations show that 3DMM and action-unit features underperform landmark distances because they over-smooth subtle facial motion, underscoring the importance of fine-grained temporal dynamics for this task. This observation directly motivates our focus on sub-landmark micro-motion and on learning motion-sensitive representations directly from raw pixels.

Pedrouzo-Rodriguez et al.~\cite{pedrouzo2025isitreally}
advance landmark-based avatar fingerprinting with a spatio-temporal graph model and broader cross-generator evaluation. Their later AVAPrintDB benchmark~\cite{pedrouzo2026avaprintdb} further shows
that current methods remain sensitive to synthesis-pipeline and domain shifts. Yet both approaches share a common structural constraint: the landmark extraction stage is frozen and non-differentiable, confining the learnable model to a pre-defined geometric vocabulary that cannot be jointly optimized for the fingerprinting objective.

Sahadewa et al.~\cite{sahadewa2026biasaverse} show that avatar fingerprinting models can exploit source-dataset shortcuts rather than true driver identity, and address this issue with a raw-RGB ResNet encoder trained using bias-averse objectives. Their approach therefore targets dataset-invariant learning through the training loss and evaluation protocol. By contrast, our method focuses on representation design: we use a micro-expression-aware F5C backbone and inter-frame feature differencing to capture fine-grained facial micro-motion while suppressing appearance structurally, without requiring explicit domain supervision.

Vahdati et al.~\cite{vahdati2025puppeteers} address avatar
fingerprinting from the latent domain, exploiting identity
leakage in the pose-and-expression codes transmitted by
AI-based videoconferencing systems. Their pose-conditioned
contrastive encoder learns persistent identity cues directly
from these latents, without reconstructing RGB frames. This
setting is complementary to ours: their method assumes
access to codec-level streaming latents, whereas we target
fingerprinting from recorded videos using raw pixels.

\subsection{Deepfake Detection via Temporal Identity Features}

Avatar fingerprinting is fundamentally different from deepfake
detection: all inputs are synthetic by design, so appearance-based
artifact cues are not informative. Nevertheless, several identity-aware
deepfake detectors are relevant in how they model temporal facial
dynamics. ID-Reveal~\cite{cozzolino2021idreveal} uses 3DMM-derived
shape and motion parameters, but its reliance on coarse parametric
features highlights the difficulty of capturing the subtle dynamics
needed for driver identification. Agarwal et al.~\cite{agarwal2019protecting}
model person-specific facial action units, but require per-identity
training and are therefore unsuitable for open-set fingerprinting.
Audio-visual inconsistency methods~\cite{cozzolino2024audiovisual,liu2024deepfake}
further assume access to reference real video, which is unavailable in
our fully synthetic setting.

More broadly, recent deepfake detection has shifted toward end-to-end
temporal analysis from raw frames. PhaseForensics~\cite{phase2025forensics}
shows that temporal phase cues carry strong forensic signal, Guo et
al.~\cite{guo2025tfcu} model temporal forgery cues across multiple
scales, and Kundu et al.~\cite{kundu2025unite} report improved
cross-domain generalization without face preprocessing. These works support our emphasis on preprocessing-free, motion-sensitive
representation learning directly from raw video.

\subsection{Facial Dynamics as an Identity Signal}

The theoretical basis for facial motion as an identity cue is
well established. Hill and Johnston~\cite{hill2001categorizing}
show that individuals can be recognized from biological face
motion alone on neutral, motion-capture-animated heads;
Knappmeyer et al.~\cite{knappmeyer2001facial} show that
personal motion signatures remain informative even when
appearance is averaged across individuals; and
O'Toole et al.~\cite{otoole2002recognizing} formalize the
dynamic identity signature hypothesis, arguing that
idiosyncratic facial motion patterns are sufficient for
recognition independently of static appearance.

This view is further supported by later work that removes
appearance altogether. ExpressionAuth~\cite{expressionauth2024}
and FacialMotionID~\cite{castro2025facialmotion} both achieve
strong identity recognition from avatar blendshapes or abstract
facial-motion signals alone. Chapariniya et
al.~\cite{chapariniya2025investigating} likewise show that
expression dynamics retain substantial identity information
after facial shape is explicitly separated. More broadly,
dynamic behavioural cues derived from whole-body keypoints
have also been shown to support person identification in
natural interaction~\cite{chapariniya2025beyond}, although
this lies outside the facial-only setting studied here.

\section{Proposed Method}
\label{sec:approach}


Let \VideoID{t}{d} denote a synthetic video produced by animating
target identity $\text{ID}_t$ with driving identity $\text{ID}_d$,
where $t$ and $d$ index the target and driver respectively.
The visual appearance of \VideoID{t}{d} is entirely determined by
$\text{ID}_t$, while the facial motion dynamics are induced by
$\text{ID}_d$.
We seek an embedding $\phi: \Video \to \mathbb{R}^{D}$, where $D$
is the embedding dimension, such that videos sharing the same
driver remain close regardless of target appearance, and videos with different drivers are well-separated even when
rendered onto the same target face:
\begin{equation}
\|\phi(\VideoID{t_1}{d}) - \phi(\VideoID{t_2}{d})\|
\;\ll\;
\|\phi(\VideoID{t}{d}) - \phi(\VideoID{t}{d'})\|,
\label{eq:objective}
\end{equation}
for all $t_1, t_2, t$ and all $d \neq d'$.
The left-hand side requires that two videos driven by the same
$\text{ID}_d$ but rendered onto different target faces $t_1$ and
$t_2$ produce similar embeddings (intra-driver distance).
The right-hand side requires that two videos rendered onto the same
target face $t$ but driven by different identities $d$ and $d'$
produce dissimilar embeddings (inter-driver distance).
Since target appearance is identical across all videos of
$\text{ID}_t$, it carries no discriminative signal for
Eq.~\ref{eq:objective}: the only information that can drive
intra-driver similarity and inter-driver separation is the
temporal facial motion of $\text{ID}_d$.
The model must therefore learn an appearance-agnostic,
driver-sensitive representation directly from raw frames.
\subsection{Architecture Overview}

\begin{figure*}[t]
\centering
\includegraphics[width=\textwidth]{final_frame_work.pdf}
\caption{\textbf{End-to-end pipeline.}
{\it Feature extraction:} 
A video clip \Video composed of $T$ grayscale frames from a generated video pass independently through the F5C backbone (ConvStack $\to$ \fcc{} $\to$ \ccc{})
to produce per-frame feature maps
$\mathbf{f}_t \in \mathbb{R}^{128\times16\times16}$.
\fcc{} provides global row/column receptive fields,
allowing each spatial location to attend across the full face;
\ccc{} models inter-regional correlations via a
$k$-NN graph ($k{=}4$), capturing coordinated muscle activations.
{\it Temporal differencing:} 
Temporal structure is introduced via inter-frame differencing
($\mathbf{d}_t = \mathbf{f}_{t+1} - \mathbf{f}_t$), which
suppresses static appearance and retains only motion dynamics.
This results in the motion tensor $\mathbf{D} \in \mathbb{R}^{128\times16\times16\times(T-1)}$.
{\it Identity encoder and training:}
The temporal identity head compresses this tensor 
$\mathbf{D}$ into the embedding $\phi(\Video)$.
The model is trained end-to-end using supervised contrastive
loss,
pulling embeddings from the same driver together 
while pushing different drivers apart.
At inference, the embedding is used to assess 
whether the video was driven by the claimed identity.
}
\label{fig:architecture}
\end{figure*}

Figure~\ref{fig:architecture} illustrates the complete model. Given a
clip of $T=64$ consecutive grayscale frames, the shared F5C backbone
produces a sequence of spatial feature maps
$\mathbf{f}_t \in \mathbb{R}^{128 \times 16 \times 16}$. Inter-frame
differences are then computed to form a motion tensor
$\mathbf{D} \in \mathbb{R}^{128 \times 16 \times 16 \times (T-1)}$,
which is finally compressed by a lightweight temporal head into a single
video-level identity embedding $\phi(V)$.
\subsection{F5C Backbone}

\mypartitle{ConvStack.}
Each frame is first mapped from
$1 \times 128 \times 128$ to $128 \times 16 \times 16$
using four convolutional layers with strides $(2,2,2,1)$, kernel sizes
$(4,3,2,1)$, BatchNorm, and ReLU. This stage provides a compact but
spatially structured representation for subsequent motion modeling.

\mypartitle{Fully-Connected Convolution (\fcc{}).}
The 128-channel feature map is split into two 64-channel halves,
each processed by a separate branch.
Meta-kernels of size~32 (cropped to the feature map spatial
extent of~16 at runtime) give each spatial location a global
receptive field across its full row and column~\cite{shao2025mol}.
Two asymmetric branches (H-then-W; W-then-H) are fused by
concatenation and $1\!\times\!1$ conv, capturing asymmetric
micro-motion (e.g., one eyebrow raised more, which is a signature of
individual motor control).

\mypartitle{Channel Correspondence Convolution (\ccc{}).}
\ccc{} models inter-regional coordination by constructing a
$k$-nearest-neighbor graph over spatial positions, where
similarity is measured in channel space via cosine distance.
Each position then aggregates differential information from
its neighbors, enabling the network to capture correlated
micro-movements across distinct facial regions, such as the
synchronized muscle activations that characterize individual
expression style, but cannot be represented by an independent
landmark trajectories.

\subsection{Inter-Frame Feature Differencing}
\label{sec:diff}

A central representation step of our framework is to compute differences between consecutive feature maps:
\begin{equation}
\mathbf{d}_t = \mathbf{f}_{t+1} - \mathbf{f}_t.
\label{eq:diff}
\end{equation}
If we assume that the feature  $\mathbf{f}_t$ can approximately 
be decomposed according to $\mathbf{f}_t = \mathbf{a} + \boldsymbol{\mu}_t$, where
$\mathbf{a}$ is target-specific appearance that varies slowly across the
clip and $\boldsymbol{\mu}_t$ is the time-varying driver-specific motion
signal. Then we have
\begin{equation}
\mathbf{d}_t
= (\mathbf{a} + \boldsymbol{\mu}_{t+1}) -
  (\mathbf{a} + \boldsymbol{\mu}_t)
= \boldsymbol{\mu}_{t+1} - \boldsymbol{\mu}_t, 
\end{equation}
which shows that temporally stable dimensions cancel algebraically,
leaving in the feature signal only the time-varying component
$\boldsymbol{\mu}_{t+1} - \boldsymbol{\mu}_t$ that encodes
driver-specific motion.
Appearance invariance therefore emerges as a structural
consequence of the differencing operation rather than as a
learned property, requiring no auxiliary objective or
architectural modification beyond the subtraction itself.

This also motivates our departure from MOL’s original frame-pair
concatenation. In micro-expression recognition, concatenation preserves expression state because the state itself is informative for the task. 
In avatar fingerprinting, however, the entanglement of the state  with target appearance can act as a distractor that is detrimental to the task.
Even if the loss and training might in theory encourage the network to remove the useless identity information, this is quite challenging and may require huge training datasets, while still not avoiding appearance shortcuts within the training stage. 
In spite of its simplicity, subtraction is therefore a much better inductive bias for enforcing sensitivity to \emph{change} rather than \emph{state}.

\subsection{Temporal Identity Head and Training}

\mypartitle{Temporal head.}
Two Conv3D layers with stride $(1,2,1)$ (temporal$\times
H\times W$) progressively downsampling $H$ from
$16\!\to\!8\!\to\!4$, leaving $W{=}16$ unchanged, 
while at the same time reducing the number of  channels
$128\!\to\!64\!\to\!32$ .
An AdaptiveAvgPool3D $(4,4,1)$ layer then equalizes the spatial
dimensions to $4{\times}4$ and collapses the temporal axis,
yielding a $32{\times}4{\times}4$ volume that is flattened to
a 512-d vector.
A two-layer MLP ($512{\to}256{\to}256$-d, ReLU, Dropout~0.3)
followed by $\ell_2$-normalisation produces the final 256-d
embedding $\phi(V)$.

\mypartitle{Supervised contrastive learning.}
Supervised contrastive learning~\cite{khosla2020supcon} is a
natural fit for avatar fingerprinting: within each batch it pulls
together all videos sharing the same driving identity across
different target appearances, generators, and recording
conditions, directly enforcing appearance invariance per gradient update. This is critical for the open-set protocol, where the embedding must generalize to identities unseen during training.
Accordingly, we train our model with the following  supervised contrastive loss:
\begin{equation}
\mathcal{L} = \sum_{i \in \mathcal{B}}
  \frac{-1}{|\mathcal{P}(i)|}
  \sum_{p \in \mathcal{P}(i)}
  \log
  \frac{
    e^{\phi_i \cdot \phi_p / \tau}
  }{
    \sum_{a \neq i} e^{\phi_i \cdot \phi_a / \tau}
  },
\end{equation}
where $\mathcal{P}(i)$ denotes all samples in the batch sharing the same
driving identity as sample $i$, and $\tau=0.07$ is the temperature.
This objective explicitly pulls together videos driven by the same
person across different targets and reenactment conditions, while
pushing away different drivers.

\mypartitle{Implementation details.}
All components are trained jointly from random initialization on
the \nvfair{} dataset, with no pretrained weights 
and no external preprocessors.
The implementation details are summarized in Table~\ref{tab:impl}.

\begin{table}[t]
\caption{Implementation details. All components are trained jointly
from scratch with no pretrained weights and no external preprocessors.}
\label{tab:impl}
\centering
\renewcommand{\arraystretch}{1.05}
\setlength{\tabcolsep}{4pt}
\begin{tabular}{ll}
\toprule
\textbf{Setting} & \textbf{Value} \\
\midrule
Input        & $T \times 1 \times 128 \times 128$ grayscale frames \\
Clip length $T$     & 64 frames ($\approx$2.6\,s at 25\,fps) \\
Embedding dim.  & 256; CCC graph $k = 4$ \\
Optimizer    & AdamW ($\beta_1{=}0.9$, $\beta_2{=}0.999$) \\
Learning rate / WD      & $10^{-3}$ cosine / $10^{-4}$; grad.\ clip 1.0 \\
Warmup       & 5 epochs linear \\
SupCon temperature & 0.07 \\
Batch        & $N{=}16$ identities $\times$ $M{=}8$ videos \\
Training     & 150 epochs $\times$ 200 steps, AMP (FP16) \\
\textbf{Total params} & \textbf{0.53\,M} (F5C: 0.06M + head: 0.47M) \\
\bottomrule
\end{tabular}
\end{table}

\section{Experimental Setup}
\label{sec:setup}

\subsection{Dataset and Task Protocol}

We evaluate our model on  \nvfair{}~\cite{prashnani2024avatar}, the first
large-scale benchmark specifically designed for avatar fingerprinting.
The dataset contains more than 650{,}000 synthetic talking-head videos
from 161 identities generated by three reenactment systems:
\fvv{}~\cite{wang2021facevidvid}, TPS~\cite{zhao2022tps}, and
LIA~\cite{wang2023lia}. The source identities are drawn from CREMA-D,
RAVDESS, and an additional videoconferencing capture collected in the
original benchmark. Each identity is associated with multiple real
videos, from which both self-reenactment and cross-reenactment videos
are synthesized.

We follow the identity-disjoint split protocol of  \nvfair{}, 112 identities for
training, 14 for validation, and 35 for testing. Crucially, the split is
\emph{identity-based}, not file-based: all videos associated with a
given identity belong to the same split, cross-reenactments are formed
only within each split, and no test identity is ever observed during
training. This setup evaluates genuine generalization to unseen drivers
rather than memorization of seen identities.

We use the original benchmark terminology. The \emph{driving identity}
is the person in the driving video, and the \emph{target identity} is
the person whose appearance is rendered in the final
video. When driving and target identities match, the result is a
\emph{self-reenactment}; when they differ, it is a
\emph{cross-reenactment}. The authentication goal is to verify that a
video of a target identity is a self-reenactment rather than an
impostor-driven cross-reenactment.

\subsection{Evaluation Protocol}

We follow the AUC computation and identity-disjoint data split
defined in Prashnani et al.~\cite{prashnani2024avatar}, using
the same 112/14/35 train/validation/test identity partition.
For each target identity, positive pairs are formed from
self-reenactment videos (driver matches target), while negative
pairs compare self-reenactments against cross-reenactments
(driver differs from target) for the same target face.
The final score is the mean AUC across all 35 test identities.

\textbf{Main evaluation (Table~\ref{tab:main}).}
The model is trained on videos from all three generators
jointly and evaluated on the combined test set, measuring
overall fingerprinting performance.

\textbf{Cross-generator evaluation (Tables~\ref{tab:gen_mol}
and~\ref{tab:gen_pdiff}).}
To assess generalization, we train three separate models,
each using videos from only one generator (\fvv{}, TPS, or LIA).
Each model is then evaluated on all three generators independently,
yielding a $3 \times 3$ matrix where rows indicate training
generator and columns indicate test generator.
Diagonal entries reflect in-domain performance; off-diagonal
entries measure cross-generator transfer.
\subsection{Experimental Conditions}

We consider three groups of controlled experiments:
input-representation conditions, clip-length ablations,
and backbone ablations.

\subsubsection{Input representation conditions}

To isolate the roles of motion, appearance, and differencing, we evaluate four F5C-based input conditions. All of them use the same F5C backbone, temporal identity head, and supervised contrastive objective; only the representation passed to the temporal head differs.

\textbf{\featdiff{} (proposed).}
Each frame is encoded independently by F5C to produce feature maps $\mathbf{f}_t$, and consecutive feature maps are subtracted as in Eq.~\ref{eq:diff}. This is our main appearance-suppressing, motion-preserving condition.

\textbf{\pdiff{}.}
Frames are differenced in pixel space before encoding, so the backbone receives $(I_{t+1} - I_t)$ rather than per-frame features. Comparing \pdiff{} with \featdiff{} isolates whether the gain comes from differencing itself or specifically from performing it in the learned feature space.

\textbf{\rawfeat{}.}
Per-frame feature maps $\mathbf{f}_t$ are passed directly to the temporal head without differencing, retaining both appearance and motion information.

\textbf{\static{}.}
A single center frame is processed, providing a purely spatial baseline with no temporal modeling.
Together, these conditions form a controlled hierarchy:
\static{} measures appearance-only capacity; \rawfeat{} adds
temporal context while retaining appearance; and \pdiff{} and \featdiff{} suppress appearance by construction, differing only in where subtraction is applied.

\subsubsection{Clip-length ablation}

To study the effect of temporal context, we vary the clip length $T \in \{16,32,64,128\}$ while keeping the rest of the pipeline fixed. All clip-length experiments use the same F5C backbone, temporal identity head, supervised contrastive loss, joint training on all three generators, and the same 150-epoch training schedule as the main model. This isolates the effect of temporal extent from other design choices.

\subsubsection{Backbone ablation}

To test whether performance depends on the motion sensitivity of the encoder rather than on differencing alone, we replace F5C with ResNet18 and evaluate matched variants under our own training protocol. This comparison is motivated by recent raw-pixel avatar fingerprinting work based on ResNet encoders, especially Sahadewa et al.~\cite{sahadewa2026biasaverse}, who use a ResNet identity encoder and bias-averse objectives to learn domain-invariant facial signatures from raw face clips. 

Our ResNet18 study is intentionally different from that method:
it is not a reproduction of their training objective, but a
controlled backbone comparison within our framework. To isolate encoder effects, we keep the data split, grayscale input format, supervised contrastive loss, embedding dimension, and 150-epoch training schedule aligned with the F5C experiments, and change only the encoder family.

For ResNet18, we evaluate practical variants that mirror the
input conditions above. In particular, \rnpdiff{} computes
frame differences in pixel space before encoding, whereas
\rnraw{} encodes raw per-frame inputs without differencing.
This setup allows us to test whether a generic appearance-oriented backbone can benefit from the same differencing strategy as F5C, or whether the gains depend on an encoder explicitly designed for fine-grained facial micro-motion.

\subsection{Temporal Sampling}
\label{sec:clip_construction}

For a chosen clip length $T$, we extract \emph{consecutive}
grayscale frames from each video rather than using strided
sampling. This is important because inter-frame differencing
operates on adjacent frames, and skipping frames would weaken
the fine-grained temporal continuity of the motion signal.

During training, we apply random temporal cropping as data
augmentation. Given a video with $N$ frames, we sample a
starting index
\begin{equation}
s \sim \mathcal{U}\bigl(0,\, \max(0, N - T)\bigr),
\end{equation}
and extract the consecutive subsequence
$\{I_s, I_{s+1}, \ldots, I_{s+T-1}\}$.
This exposes the model to different temporal segments across
epochs and reduces sensitivity to any particular clip position.

During evaluation, we use deterministic center cropping for
reproducibility:
\begin{equation}
s = \left\lfloor \frac{N - T}{2} \right\rfloor.
\end{equation}
The clip is then formed as
$\{I_s, I_{s+1}, \ldots, I_{s+T-1}\}$.
Videos shorter than $T$ frames are padded by repeating the
final frame until the required length is reached.
\section{Results}
\label{sec:results}

\subsection{Method Comparison}

Table~\ref{tab:method_comparison} compares our approach with
prior avatar fingerprinting systems from a deployment
perspective. Landmark-based methods~\cite{prashnani2024avatar,pedrouzo2025isitreally}
require an external per-frame keypoint extraction stage before
the fingerprinting model can operate. In contrast, our method
is trained end-to-end from raw frames and removes this
landmark-tracking dependency entirely. Despite this simpler
pipeline, our best model remains competitive with prior
landmark-based systems while preserving cross-generator
generalization.

\begin{table*}[t]
\caption{Practical deployment properties of avatar fingerprinting
methods.
\yes{} = yes; \no{} = no.
\textbf{E2E}: end-to-end trainable.
\textbf{KP-free}: no external keypoint detector.
\textbf{RT}: no per-frame landmark-tracking overhead.
$^\dagger$Per-generator AUC from respective papers.
$^\ddagger$Mean AUC across generators, our experiments.}
\label{tab:method_comparison}
\centering
\renewcommand{\arraystretch}{1.14}
\setlength{\tabcolsep}{4.5pt}
\begin{tabular}{lcccccccc}
\toprule
\textbf{Method} & \textbf{Dataset}
  & \textbf{Input type}
  & \textbf{E2E} & \textbf{KP-free} & \textbf{RT}
  & \textbf{AUC}$^\dagger$ & \textbf{Cross-gen} \\
\midrule
NVFAIR~\cite{prashnani2024avatar}
  & \nvfair{}
  & Landmark dist.\ (126~pts)
  & \no & \no & \no & 0.87 & \yes \\
IsItReallyYou~\cite{pedrouzo2025isitreally}
  & GAGAvatar-BM
  & Landmark feat.\ (109~pts)
  & \no & \no & \no & 0.80 & --- \\
Unmasking Puppeteers~\cite{vahdati2025puppeteers}
  & NVIDIA-VC 
  & Codec latents
  & \no & \yes & \yes & --- & \yes \\
\midrule
ResNet18 + \pdiff{}
  & \nvfair{} & Raw grayscale
  & \yes & \yes & \yes & 0.759$^\ddagger$ & \no\\
F5C + \static{} 
  & \nvfair{} & Raw grayscale
  & \yes & \yes & \yes & 0.610 & \no \\
F5C + \rawfeat{} 
  & \nvfair{} & Raw grayscale
  & \yes & \yes & \yes & 0.649 & \no \\
F5C + \pdiff{} 
  & \nvfair{} & Raw grayscale
  & \yes & \yes & \yes & 0.861 & \yes \\
F5C + \featdiff{} \textbf{(ours)}
  & \nvfair{} & Raw grayscale
  & \yes & \yes & \yes
  & \textbf{0.877}$^\ddagger$ & \yes \\
\bottomrule
\end{tabular}
\end{table*}

\subsection{Main Benchmark Performance}

Table~\ref{tab:main} reports performance on the full
\nvfair{} benchmark across all three generators. Our proposed
\featdiff{} model achieves an overall AUC of 0.877,
outperforming the landmark-based NVFAIR baseline
(0.853) by $+0.024$. At the per-generator level,
\featdiff{} is essentially on par with the baseline on
\fvv{} (0.869 vs.\ 0.870), while improving substantially on
LIA ($+0.042$) and TPS ($+0.030$). These gains are obtained
without any external landmark extraction or preprocessing.

\pdiff{} also performs competitively, reaching 0.861 overall
AUC. This shows that differencing itself is a strong design
principle. However, \featdiff{} consistently outperforms
\pdiff{}, indicating that applying differencing in the learned
feature space yields more discriminative motion
representations than applying it directly in pixel space.

\begin{table}[t]
\caption{Per-generator AUC on \nvfair{} over 35 test identities.
$^\dagger$Reported in~\cite{prashnani2024avatar}.}
\label{tab:main}
\centering
\renewcommand{\arraystretch}{1.05}
\begin{tabular}{lcccc}
\toprule
\textbf{Method} & \textbf{Overall} & \textbf{FV2V} & \textbf{LIA} & \textbf{TPS} \\
\midrule
NVFAIR$^\dagger$ & 0.853 & \textbf{0.870} & 0.840 & 0.850 \\
\midrule
\featdiff{} (ours) & \textbf{0.877} & 0.869 & \textbf{0.882} & \textbf{0.880} \\
\pdiff{} (ours)    & 0.861 & 0.849 & 0.866 & 0.868 \\
\bottomrule
\end{tabular}
\end{table}

\subsection{Cross-Generator Generalization}

Tables~\ref{tab:gen_mol} and~\ref{tab:gen_pdiff} summarize
cross-generator transfer. Both \featdiff{} and \pdiff{}
generalize well across unseen generators, with positive
average gains on two of the three test generators relative to
the landmark baseline. The strongest single result is obtained
when training and testing on LIA, where \featdiff{} reaches
0.902 AUC and \pdiff{} reaches 0.893.

The similar pattern across the two tables suggests that the
differencing operation is the main source of cross-generator
robustness. At the same time, \featdiff{} remains consistently
stronger overall, showing that motion-sensitive feature-space
representations provide an additional advantage beyond
differencing alone. The only setting in which our method
falls below the landmark baseline is when training on
\fvv{} and testing on TPS or LIA, which mirrors an already
weak transfer pattern in the NVFAIR baseline and therefore
appears to reflect generator mismatch rather than a limitation
specific to our architecture.

\begin{figure*}[t]
\centering
\includegraphics[width=0.9\textwidth,height=0.2\textheight]{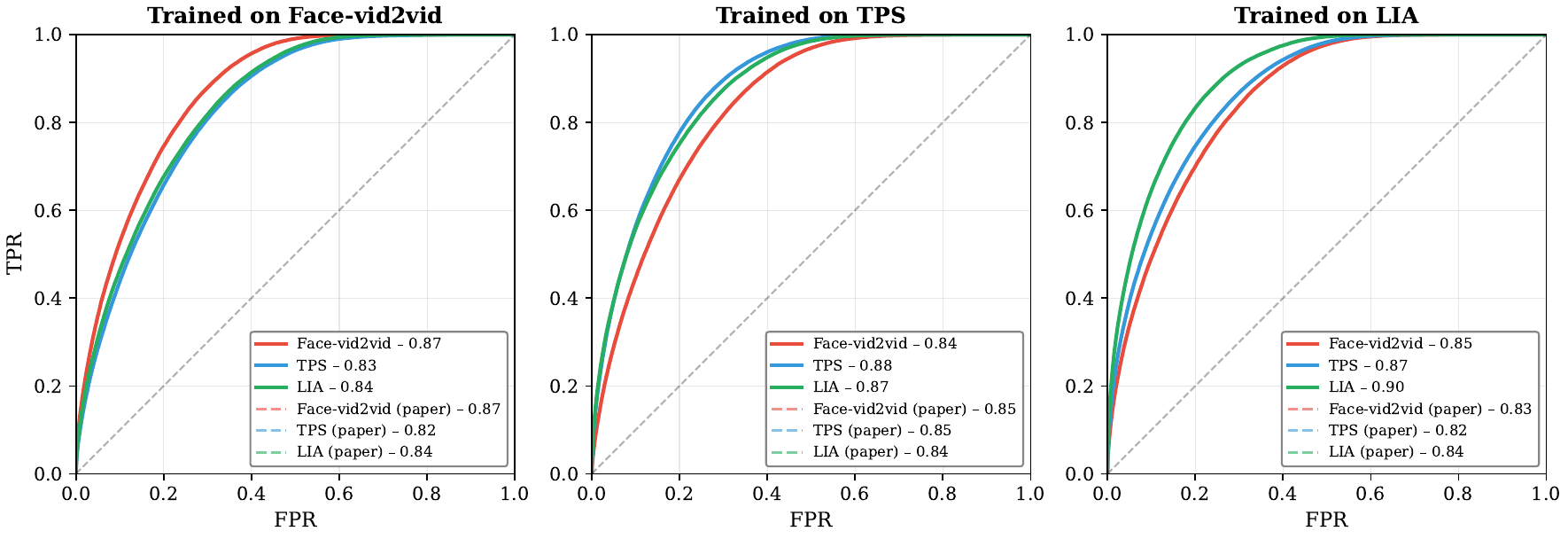}
\caption{Cross-generator generalization for \featdiff{} (solid) versus the
NVFAIR baseline (dashed). Each panel corresponds to a model trained on a
single generator and evaluated on all three generators.}
\label{fig:gen_mol}
\end{figure*}

\begin{table}[t]
\caption{\featdiff{} cross-generator AUC. Diagonal entries are in
\textbf{bold}.}
\label{tab:gen_mol}
\centering
\renewcommand{\arraystretch}{1.0}
\setlength{\tabcolsep}{5pt}
\begin{tabular}{lccc}
\toprule
\textbf{Train $\backslash$ Test} & \textbf{FV2V} & \textbf{TPS} & \textbf{LIA} \\
\midrule
\fvv{} & \textbf{0.869} & 0.833 & 0.842 \\
TPS    & 0.838 & \textbf{0.879} & 0.872 \\
LIA    & 0.852 & 0.869 & \textbf{0.902} \\
\bottomrule
\end{tabular}
\end{table}

\begin{table}[t]
\caption{\pdiff{} cross-generator AUC.}
\label{tab:gen_pdiff}
\centering
\renewcommand{\arraystretch}{1.0}
\setlength{\tabcolsep}{5pt}
\begin{tabular}{lccc}
\toprule
\textbf{Train $\backslash$ Test} & \textbf{FV2V} & \textbf{TPS} & \textbf{LIA} \\
\midrule
\fvv{} & \textbf{0.865} & 0.822 & 0.816 \\
TPS    & 0.839 & \textbf{0.891} & 0.873 \\
LIA    & 0.857 & 0.862 & \textbf{0.893} \\
\bottomrule
\end{tabular}
\end{table}

\begin{table}[t]
\caption{AUC across four F5C-based representation conditions at
$T{=}64$.}
\label{tab:ablation}
\centering
\renewcommand{\arraystretch}{1.05}
\setlength{\tabcolsep}{3pt}
\begin{tabular}{lcccc}
\toprule
\textbf{Condition} & \textbf{Overall} & \textbf{FV2V} & \textbf{TPS} & \textbf{LIA} \\
\midrule
\featdiff{} & \textbf{0.877} & \textbf{0.869} & \textbf{0.882} & \textbf{0.880} \\
\pdiff{}    & 0.861 & 0.849 & 0.866 & 0.868 \\
\rawfeat{}  & 0.649 & 0.719 & 0.684 & 0.682 \\
\static{}   & 0.610 & 0.685 & 0.631 & 0.630 \\
\bottomrule
\end{tabular}
\end{table}
\begin{table}[t]
\caption{Clip-length ablation (F5C + \featdiff{}, SupCon, 150 epochs,
all generators jointly). $T{=}64$ is the default configuration.}
\label{tab:clip_length}
\centering
\renewcommand{\arraystretch}{1.05}
\setlength{\tabcolsep}{4pt}
\begin{tabular}{ccccc}
\toprule
$T$ & \textbf{Overall} & \textbf{FV2V} & \textbf{TPS} & \textbf{LIA} \\
\midrule
16  & 0.813 & 0.808 & 0.817 & 0.816 \\
32  & 0.862 & 0.853 & 0.868 & 0.866 \\
64  & 0.877 & 0.869 & 0.880 & 0.882 \\
\textbf{128} & \textbf{0.891} & \textbf{0.883} & \textbf{0.895} & \textbf{0.896} \\
\bottomrule
\end{tabular}
\end{table}
\begin{table}[t]
\caption{Backbone ablation: ResNet18 variants vs.\ F5C (\featdiff{}).
All conditions use SupCon ($\tau{=}0.07$), $T{=}64$, 150 epochs,
and are trained jointly on all generators.}
\label{tab:backbone}
\centering
\renewcommand{\arraystretch}{1.05}
\setlength{\tabcolsep}{3pt}
\begin{tabular}{lcccc}
\toprule
\textbf{Method} & \textbf{Overall} & \textbf{FV2V} & \textbf{TPS} & \textbf{LIA} \\
\midrule
\rnstatic{}  & 0.581 & 0.695 & 0.596 & 0.599 \\
\rnraw{}     & 0.581 & 0.693 & 0.609 & 0.608 \\
\rndiff{}    & 0.547 & 0.567 & 0.553 & 0.556 \\
\rnpdiff{}   & \textbf{0.759} & \textbf{0.785} & \textbf{0.763} & \textbf{0.764} \\
\midrule
\featdiff{} (F5C, ours) & \textbf{0.877} & \textbf{0.869} & \textbf{0.880} & \textbf{0.882} \\
\bottomrule
\end{tabular}
\end{table}
\subsection{Controlled Ablation Analyses}

\subsubsection{Representation Ablation}

Table~\ref{tab:ablation} isolates the roles of motion,
appearance, and differencing. \featdiff{} performs best,
while \static{} performs worst, showing that temporal motion
is essential for driver identification. \rawfeat{} also
underperforms substantially, indicating that preserving
appearance harms identity separation rather than helping it.
Since \rawfeat{} contains both appearance and motion whereas
\featdiff{} removes only the temporally stable component, this
gap suggests that appearance-correlated dimensions introduce
conflicting gradients during training.

The comparison between \featdiff{} and \pdiff{} shows that
differencing is the main source of discriminative power, but
that feature-space differencing remains preferable when the
backbone preserves useful motion structure.
\subsubsection{Temporal Context Ablation}

Table~\ref{tab:clip_length} shows that performance improves
monotonically as the clip length increases from 16 to 128
frames. This confirms that longer temporal context provides
additional identity-bearing motion information. We use
$T{=}64$ as the default configuration because it offers a good
trade-off between accuracy, memory, and runtime while still
capturing approximately 2.6\,s of consecutive micro-motion.

\subsubsection{Backbone Ablation}

To test whether the gains arise from differencing alone or
from the motion sensitivity of the encoder, we replace F5C
with ResNet18 and re-evaluate the same four conditions under
matched training settings. Table~\ref{tab:backbone} shows
that the best ResNet18 variant remains far below F5C despite
using substantially more parameters (11.6\,M vs.\ 0.53\,M).

In particular, ResNet18 with feature-space differencing
collapses below even the static baseline, indicating that relying on a generic appearance-oriented encoder does not preserve useful
inter-frame variation. By contrast, the \fcc{} and \ccc{}
modules in F5C are specifically designed to encode fine-scale,
sub-landmark motion patterns that vary measurably across
frames even when global appearance remains constant.
Pixel-space differencing partially recovers performance by
making motion explicit at the input, but it still remains well
below F5C with feature-space differencing. This confirms that
effective avatar fingerprinting depends on both differencing
and a backbone explicitly suited to micro-motion encoding.

\subsection{Limitations and Future Work}

Our evaluation is currently limited to NVFAIR, which covers
three pre-diffusion reenactment generators. Although the
proposed method shows strong overall and cross-generator
performance in this setting, its behavior on newer
diffusion-based portrait generators and on subjects with very
weak facial expressivity remains to be established. A natural
next step is evaluation on broader benchmarks such as
AVAPrintDB and GAGAvatar-based test sets, which would
better assess robustness under contemporary synthesis
pipelines and more diverse motion regimes.
Several methodological limitations also remain. First, we do
not study robustness against adaptive or adversarial attacks;
an informed attacker may attempt to suppress or perturb the
temporal cues exploited by inter-frame differencing. Second,
while the appearance-suppression effect is supported by both
the ablation results and the algebraic interpretation in
Eq.~\ref{eq:diff}, a more formal disentanglement analysis
would strengthen the theoretical grounding of the method.
Finally, extensions such as multi-clip score aggregation and
joint training across generators may further improve
verification stability and transfer.
From a deployment perspective, avatar fingerprinting should be viewed
as one component of a broader trust pipeline, requiring continued
evaluation under stronger attack models and careful governance of
biometric embeddings.

\section{Conclusion}
\label{sec:conclusion}

We presented a preprocessing-free avatar fingerprinting framework
that combines a micro-expression-aware F5C backbone with inter-frame
feature differencing, suppressing target appearance by construction
while retaining driver-specific micro-motion.
On \nvfair{}, our 0.53\,M-parameter model achieves 0.877 overall AUC,
matches or exceeds the landmark-based baseline on most cross-generator
pairs, and requires no external landmark tracking.
Controlled ablations establish the key design principles: temporal
differencing is the dominant source of identity signal, raw appearance
features are actively detrimental, and feature-space differencing
requires a backbone explicitly sensitive to fine-grained facial
micro-motion.
These findings position inter-frame feature differencing as a simple
but principled mechanism for appearance-agnostic driver verification,
pointing avatar fingerprinting beyond fixed geometric pipelines toward
more robust verification under diverse talking-head generation systems.

{\small
\balance
\bibliographystyle{IEEEtran}
\bibliography{references_fg}
}

\end{document}